\documentclass[11pt]{article}
\usepackage{fullpage}
\usepackage{amsmath,amssymb}
\usepackage{array}
\usepackage{dsfont}
\usepackage{makeidx}
\usepackage{hyperref}
\usepackage{url}

\usepackage{algorithm}
\usepackage{algorithmic}
\usepackage{amsmath}
\usepackage{rotating}

\usepackage{graphicx}

\def\inner#1#2{ {\langle #1, #2 \rangle} }

\title{Statistical exponential families: A digest with flash cards\footnote{See the {\tt jMEF} library, a Java package for processing mixture of exponential families. Available for download at {\tt http://www.lix.polytechnique.fr/\~{}nielsen/MEF/}}}
\author{Frank Nielsen\footnote{\'Ecole Polytechnique (France) and Sony Computer Science Laboratories Inc. (Japan).} and Vincent Garcia\footnote{\'Ecole Polytechnique (France).}}
\date{\today}
\makeindex

\begin{document}


\renewcommand{\arraystretch}{1.8}


\date{\today{} (v2.0)}
\maketitle

\begin{abstract}
This document describes concisely the ubiquitous class of exponential family distributions met in statistics.
The first part recalls definitions and summarizes main properties and duality with Bregman divergences (all proofs are skipped).
The second part lists decompositions and related formula of common exponential family distributions.
We recall the Fisher-Rao-Riemannian geometries and the dual affine connection information geometries of statistical manifolds.
It is intended to maintain and update this document and catalog by adding new distribution items.
\end{abstract}

\newpage
\part{A digest of exponential families}
\def\X{\mathcal{X}}
\def\P{\mathcal{P}}
\def\dim{\mathrm{dim}}
\def\grad{\nabla}
\def\dt{\mathrm{d}t}
\def\dx{\mathrm{d}x}
\def\ds{\mathrm{d}s}

\def\dtheta{\mathrm{d}\theta}
\def\KL{\mathrm{KL}}
\def\var{\mathrm{var}}
\def\argmax{\mathrm{argmax}}
\def\Pr{\mathrm{Pr}}
\def\Tr{\mathrm{Tr}}
\def\dom{\mathrm{dom}}
\def\M{\mathcal{M}}
\def\W{\mathcal{W}}

\def\ADD#1{\fbox{#1}}

\section{Essentials of exponential families}

\subsection{Sufficient statistics}

\begin{figure}[h]
\centering
\includegraphics[width=0.7\textwidth]{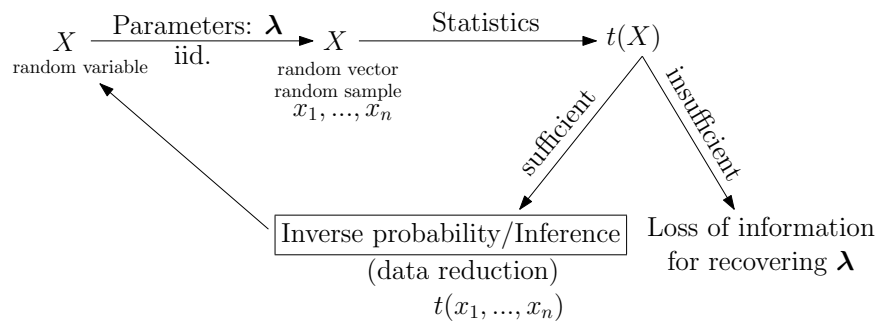}

\caption{Collecting statistics of a parametric random vector allows one to perform data reduction for inference problem if and only if those statistics are sufficient.
Otherwise,  loss of information occurs and the parameters of the family of distributions cannot be fully recovered from the (insufficient) statistics.
\label{stat:fig:sufficientstat}}
\end{figure}

A fundamental problem in statistics is to recover the  model parameters $\lambda$  from a given  set of observations $x_1, ...,$ etc.\index{model parameter}
Those samples are assumed to be  randomly drawn from an independent and identically-distributed random vector with associated density $p(x;\lambda)$.\index{random vector}
Since the sample set is finite, statisticians estimate a close approximation $\hat\lambda$ of the true parameter.
However, a surprising fact is that one can collect and concentrate from a random sample  all  necessary  information for recovering/estimating the parameters.
The information is collected into  a few elementary statistics of the random vector, called the sufficient statistic.\footnote{First coined by statistician Sir Ronald Fisher in 1922.}\index{sufficient statistic}
Figure~\ref{stat:fig:sufficientstat} illustrates the notions of statistics and sufficiency.\index{sufficiency}

It is challenging to find  sufficient statistics for a given  parametric probability distribution function $p(x;\lambda)$.
The Fisher-Neyman factorization theorem\index{Fisher-Neyman factorization}  allows one to easily identify those sufficient statistics from the decomposition characteristics of the probability distribution function.
A statistic $t(x)$ is sufficient if and only if the density can be decomposed as

\begin{equation}
p(x;\lambda)=a(x)b_\lambda(t(x)),
\end{equation}
where $a(x)\geq 0$ is a non-negative function independent of the distribution parameters. 
The class of exponential families encompass most common statistical distributions and are provably the only ones (under mild conditions) that allow one for data reduction.

\subsection{Exponential families: Definition and properties}
An exponential family is a set of probability distributions  admitting the following canonical decomposition:
\begin{equation}
	p(x; \theta) = \exp \left( \langle t(x), \theta \rangle - F(\theta) + k(x) \right)
\end{equation}
where
\begin{itemize}
	\item $t(x)$ is the sufficient statistic\index{sufficient statistic},
	\item $\theta$ are the natural parameters\index{natural parameter},
	\item $\langle .,. \rangle$ is the inner product\index{inner product}\index{dot product} (commonly called dot product\index{dot product}),
	\item $F(\cdot)$ is the log-normalizer\index{log-normalizer},
	\item $k(x)$ is the carrier measure\index{carrier measure}. 
\end{itemize}

The exponential distribution is said univariate if the dimension of the observation space $\X$ is 1D, otherwise it is said multivariate.
The order\index{order} $D$ of the family is the dimension of the natural parameter space $\P_\Theta$.
Part~II reports the canonical decompositions of common exponential families.
Note that handling probability measures allows one to consider both probability densities and probability mass functions in a common framework. 
Consider $(X,a,\mu)$ a measurable space (with $a$ a $\sigma$-algebra) and $f$ a
measurable map, the probability measure is defined as
$P_\theta(\dx)=p_F(x;\theta)\mu(\dx)$.
$a$ is often a $\sigma$-algebra on the Borel sets with $\mu$ the Lebesgue measure
restricted to $X$.

For example, 
\begin{itemize}
\item Poisson distributions are univariate exponential distributions of order $1$ (e.g., $\dim\X=1$ and $\dim\P=1$) with associated probability mass function:
\begin{equation}
\Pr(x=k; \lambda)=\frac{\lambda^k e^{-\lambda}}{k!},
\end{equation} 
for $k\in\mathbb{N}$.

The canonical exponential family decomposition yields:
\begin{itemize}
	\item $t(x)=x$ is the sufficient statistic,
	\item $\theta=\log\lambda$ are the natural parameters,
	\item $F(\theta)=\exp\theta$ is the log-normalizer,
	\item $k(x)=-\log x!$ is the carrier measure. 
\end{itemize}

\item 1D Gaussian distributions are univariate distributions of order $2$ (e.g., $\dim\X=1$ and $\dim\P=2$),  characterized by two parameters $(\mu,\sigma)$ with associated density 
\begin{equation}
\frac{1}{\sigma\sqrt{2\pi}} e^{-\frac{1}{2}\left(\frac{x-\mu}{\sigma}\right)^2},
\end{equation}
for $x\in\mathbb{R}$.

The canonical exponential family decomposition yields:
\begin{itemize}
	\item $t(x)=(x,x^2)$ is the sufficient statistic,
	\item $\theta=(\theta_1,\theta_2)=(\frac{\mu}{\sigma^2},-\frac{1}{2\sigma^2})$ are the natural parameters,
	\item $F(\theta)=-\frac{\theta_1^2}{4\theta_2} + \frac{1}{2} \log \left( -\frac{\pi}{\theta_2} \right)$ is the log-normalizer,
	\item $k(x)=0$ is the carrier measure. 
\end{itemize}

\end{itemize}

Exponential families~\cite{brown-1986} are characterized by their strictly convex and differentiable functions $F$, called log-normalizer
 (or cumulant\index{cumulant function}/partition\index{partition} function).
 The sufficient statistics $t(x): \X \mapsto \P_\Theta$ is said minimal\index{sufficient statistic!minimal} if they are affinely independent.
 The carrier measure is usually the Lebesgue (e.g., Gaussian, Rayleigh, etc.) or counting measures (e.g., Poisson, binomial, etc.).
 Note that 
 \begin{equation}
 F(\theta)=\log \int_x \exp (<t(x),\theta>+k(x)) \dx.
 \end{equation}
 It is thus easy to build an exponential family: fix $k(x)=0$ and let us choose for $t(x)$ an arbitrary function for a given domain $x\in [x_{\mathrm{min}},x_{\mathrm{max}}]$.
 For example, consider $t(x)=x$ for $x\in[-\infty,1]$, then $F(\theta)=\int_x \exp \theta x \dx=[\frac{e^{\theta x}}{\theta}]_{x=-\infty}^1=\frac{1}{\theta}(e^\theta-1)$.

By remapping $t(x)$ to $y$, we can consider without loss of generality regular exponential family\index{regular} where the dimension of the observation space matches the parameter space.  The regular canonical decomposition of the density simplifies to
\begin{equation}
	p(y;\theta) = \exp \left( \langle y, \theta \rangle - F(\theta)\right)
\end{equation}
with respect to the base measure $h(x)=\exp(k(x))$.

\noindent Exponential families include many familiar distributions~\cite{brown-1986}, characterized by their log-normaliser functions:\\
{\it
Gaussian or normal (generic, isotropic Gaussian, diagonal Gaussian, rectified Gaussian or Wald distributions, log-normal), Poisson, Bernoulli, binomial, multinomial (trinomial, Hardy-Weinberg distribution), Laplacian, Gamma (including the chi-squared), Beta, exponential, Wishart, Dirichlet, Rayleigh, probability simplex, negative binomial distribution, Weibull, Fisher-von Mises, Pareto distributions, skew logistic, hyperbolic secant, negative binomial, etc.
}

However, note that the uniform distribution\index{uniform distribution} does not belong to the exponential families.
\begin{sidewaystable}
\centering
\includegraphics[width=\textwidth]{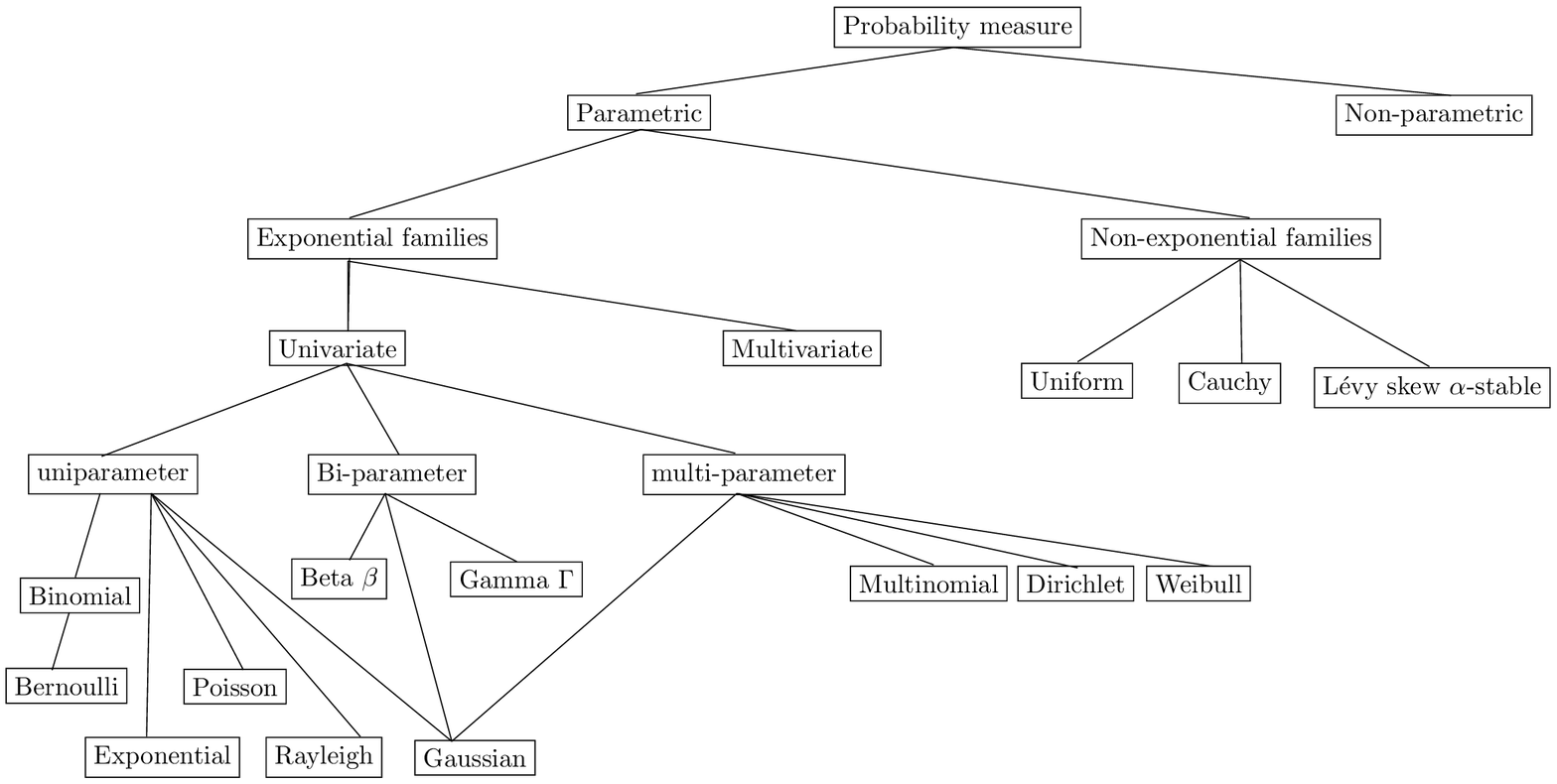}
\caption{Partial taxonomy of statistical distributions. 
\label{expfam:class}}
\end{sidewaystable}

The observation/sample space $\cal{X}$ can be of different types like integer (e.g., Poisson), scalar (e.g., normal), categorical (e.g., multinomial), symmetric positive definite matrix (e.g., Wishart), etc. 
For the latter case, the inner product of symmetric positive definite matrices is defined as the matrix trace of the product $<X,Y>=\Tr(XY)$, the sum of the eigenvalues of the matrix product $X\times Y$.

The $k$-order non-centered moment\index{moment} is defined as the expectation $E[X^k]$.
The first moment is called the mean\index{mean} $\mu$ and the second centered moment $E[(X-\mu)^T(X-\mu)]$ is called the variance-covariance or dispersion matrix.\index{variance-covariance matrix}\index{dispersion matrix}

For exponential families, we have 
\begin{equation}
E[t(X)]=\mu=\grad F(\theta)
\end{equation}

\begin{equation}
E[(t(X)-\mu)^T(t(X)-\mu)]=\grad^2 F(\theta) 
\end{equation}
Notation $\grad^2 F$ denotes the Hessian of the log-normalizer. It is a positive definite matrix since $F$ is strictly convex and differentiable function.
In fact, exponential families have all finite moments, and $F$ is $C^{\infty}$ differentiable.
Thus Cauchy distributions are provably not an exponential family, since it has no defined mean.
Another widely used  family of distributions that are not exponential families are the L\'evy skew $\alpha$-stable distributions.

In practice, we need to consider minimal exponential family with the sufficient statistics $t(x)$.
Since they are several ways to decompose a density/probability mass according to the terms ${\theta},t(x),F(\theta)$ and $k(x)$, we adopt the following  basic conventions:

The sufficient statistic $t(x)$   should be elementary functions (often polynomial functions) with leading unit coefficient.
The carrier measure should not have constant term, so that we prefer to absorb the constant in the log-normalizer.
Finally, we denote by  ${\Lambda}$ the traditional source parameters, and by $\P_{\Lambda}$ the source parameter space (canonical parameter space\index{canonical parameter}).

The natural parameter space 
\begin{equation}
\P_\Theta=\{\Theta\ |\ |F(\Theta)|<+\infty \}
\end{equation} is necessarily an open convex set.
Although the product of exponential families is an (unnormalized) exponential family, the mixture of exponential families is not an exponential family.

The moment generating function\index{moment generating function} of an exponential family $\{p_F(x;\theta)\ |\ \theta\in\P_{\Theta}\}$ is:
\begin{equation}
m_\theta(x)= \exp \left( F(\theta+x)-F(\theta) \right)
\end{equation}
Function $F$ is thus sometimes called logarithmic moment generating function~\cite{amari-2000} (p. 69).\index{logarithmic moment generating function}

The cumulant generating function\index{cumulant generating function} is defined as 
\begin{equation}
\kappa_\theta(x)=\log m_\theta(x)=F(\theta+x)-F(\theta)
\end{equation}

Exponential families can be generated from Laplace transforms~\cite{eflaplace-2007}.\index{transform!Laplace}
Let $H(x)$ be a bounded non-negative measure\index{measure}  with density $h(x)=\exp k(x)$.
Consider the Laplace transform:

\begin{equation}
L(\theta)=\int \exp(<x,\theta>)h(x) \dx
\end{equation}

Since $L(\theta)>0$, we can rewrite the former equation to show that

\begin{equation}
p(x;\theta)=\exp \left( <x,\theta> - \log L(\theta) \right)
\end{equation}
is a probability density with respect to the measure $H(x)$.
That is, the log-normalizer of the exponential family is the logarithm of the Laplace transform of the measure $H(x)$.

\subsection{Dual parameterizations: Natural and expectation parameters}
\index{parameter!natural}\index{parameter!expectation}\index{parameter!moment}

A fundamental duality of convex analysis is the Legendre-Fenchel transform.\index{transform!Legendre-Fenchel}
Informally, it says that strictly convex and differentiable functions come by pairs.

For the log-normalize $F$, consider its Legendre dual $G=F^*$ defined by the ``slope'' transform

\begin{equation}
G(\eta)= \sup_{\theta\in \P_{\Theta}} <\theta,\eta>-F(\theta).
\end{equation}

The extremum is obtained for $\eta=\nabla F(\theta)$.
$\eta$ is called the moment parameter, since for exponential families we have $\eta=E[t(X)]=\mu=\grad F(\theta)$.
Gradients of conjugate pairs are inversely reciprocal $\nabla F^*=(\nabla F)^{-1}$, and therefore  $F^*=\int \nabla F^*=\int(\nabla F)^{-1}$.

\begin{figure}
\centering

\includegraphics[width=0.6\textwidth]{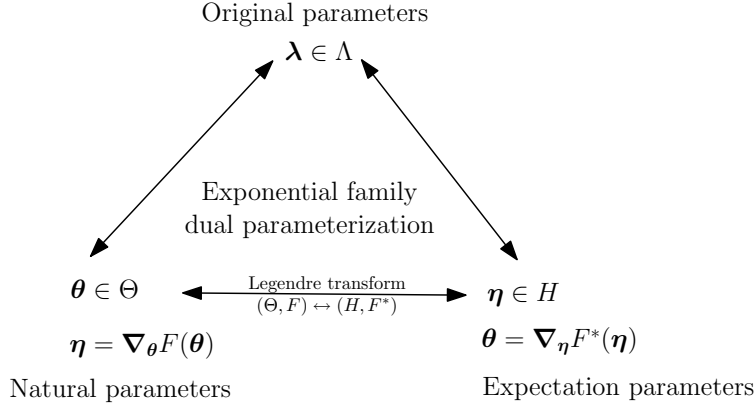}

\caption{Dual parameterizations of exponential families from Legendre transformation.
\label{expfam:fig:dualparam} }
\end{figure}

Thus to describe a member of an exponential family, we can either use the source or canonical natural/expectation parameters.
Figure~\ref{expfam:fig:dualparam} illustrates the conversions procedures.

\subsection{Geometry of exponential families: Riemannian and information geometries}
A family of parametric distributions $\{p(x;\theta)\}$ (exponential or not) may be thought as a smooth manifold that can be studied  using the framework of differential geometry~\cite{Lauritzen:1987}. We review two main types of geometries: (1) Riemannian geometry defined by a bilinear tensor with an induced Levi-Cevita connection, and non-metric geometry induced by a symmetric affine connection. 

Cencov\footnote{also written as Chentsov} proved~\cite{Cencov-1972} (see also~\cite{lebanon-05} and~\cite{grasselli-2001} for an equivalent in quantum information geometry)  that the only Riemannian metric that ``makes  sense'' for statistical manifolds is the Fisher information metric:

\begin{equation}
I(\theta)=\left[ 
\int \frac{\partial\log p(x;\theta)}{\partial\theta_i} \frac{\partial\log p(x;\theta)}{\partial\theta_j} p(x;\theta) \dx
\right]=[g_{ij}]
\end{equation}

The infinitesimal length element is given by 
\begin{equation}
\ds^2=\sum_{i=1}^d \sum_{i=1}^d \dtheta_i^T \nabla^2 F(\theta) \dtheta_j
\end{equation}

Cencov proved that for a non-singular transformation of the parameters $\lambda=f(\theta)$, the information matrix 
\begin{equation}
I(\lambda)=\left[\frac{\partial\theta_i}{\partial\lambda_j}\right] I(\theta)  \left[\frac{\partial\theta_i}{\partial\lambda_j}\right],
\end{equation}
is such that $\ds^2(\lambda)=\ds^2(\theta)$. 
Equipped with the tensor $I(\theta)$, the metric distance between two distributions on a statistical manifold  can be computed from the geodesic length (e.g., shortest path):

\begin{equation}
D(p(x;\theta_1),p(x;\theta_2))=\min_{\theta(t)\ |\ \theta(0)=\theta_1,\theta(1)=\theta_2}
\int_0^1\sqrt{ \left(\frac{\dtheta}{\dt}\right)^T I(\theta)  \frac{\dtheta}{\dt} }\dt
\end{equation}
Rao's geodesic distance is invariant by non-singular transformations.\index{Rao's distance}
The multinomial Fisher-Rao-Riemannian geometry yields a spherical geometry, and the normal Fisher-Rao-Riemannian geometry yields a hyperbolic geometry~\cite{kass:1997,fishernormal-2005}.
Indeed, the Fisher information matrix for univariate normal distributions is
\begin{equation}
I(\theta)=\frac{1}{\sigma^2} \begin{pmatrix}1 & 0\cr 0 & 2\end{pmatrix}.
\end{equation}

The Fisher information matrix can be interpreted as the Hessian of the Shannon entropy:

\begin{equation}
g_{ij}(\theta)=-E\left[\frac{\partial^2 \log p(x;\theta)}{\partial\theta_i\partial\theta_j} \right]=\frac{\partial^2H(p)}{\partial\theta_i\partial\theta_j},
\end{equation} 
 with $H(p)=-\int p(x;\theta)\log p(x;\theta) \dx$.

For an exponential family, the Kullback-Leibler divergence is a Bregman divergence on the natural parameters.
Using Taylor approximation with exact remainder, we get
$\KL(\theta || \theta+\dtheta)=\frac{1}{2}\dtheta^T \nabla^2F(\theta)\dtheta$.
Moreover, the infinitesimal Rao distance is $\sqrt{\dtheta^T I(\theta)\dtheta}$
for $I(\theta)=\nabla^2F(\theta)$. We deduce that
$D(\theta,\theta+\dtheta)=\sqrt{2\KL(\theta||\theta+\dtheta)}$.

The inner product\footnote{Technically speaking, it is a bilinear symmetric positive definite operator. A tensor $[T_p]_2^2$ of covariant degree $2$ and of contravariant degree $0$: $<\cdot,\cdot>_p: T_p\times T_p \rightarrow \mathbb{R}$.} of two vectors $x$ and $y$ at a tangent space of point $p$ is
\begin{equation}
<x,y>_p= x^T g_p y .
\end{equation}
The length of a vector $v$ in the tangent space at $T_p$ at $p$ is defined by $\|v \|_p=\sqrt{<v,v>_p}$.

For exponential families, the logarithm of the density is concave (since $F$ is convex), and we have 

\begin{equation}
I(\theta)=\left[\frac{\partial\eta}{\partial\theta}\right]=\nabla^2 F(\theta)=I^{-1}(\eta)=\left[\frac{\partial\theta}{\partial\eta}\right]
\end{equation}

That is, the Fisher information is the Hessian of the log-normalizer $\grad^2 F(\theta)$. 

To a given Riemannian manifold $\M$ with tensor $G$, we may associate\footnote{Or view/interpret the manifold as a statistical manifold.} a probability measure as follows: Let $p(\theta)=\frac{1}{V}\sqrt{\det G(\theta)}$ with overall volume $V=\int_{\theta\in\Theta} \sqrt{G(\theta)} \dtheta$.
These distributions are built from infinitesimal volume element and bear the names of Jeffreys priors.\index{Jeffreys prior}
They are commonly used in Bayesian estimation~\cite{amari-2000,kass:1997}.

Furthermore, in an exponential family manifold, the geometry is flat and  $\theta/\eta$ are dual coordinate systems.

Amari~\cite{amari-2000} focused on a pair of dual affine mixture/exponential connections $\nabla^{m}$ and $\nabla^{e}$ induced by a contrast function $F$ (also called potential function).\index{potential function}\index{contrast function}

A connection $\nabla$ yields a function $\prod_{p,q}$  that maps vectors in any pair of tangent spaces $T_p$ and $T_q$. 
An affine connection is defined by $d^3$ coefficients. Amari's investigated thoroughly the $\alpha$-connections and showed
that $\nabla^{(0)}$, the metric Levi-Civita connection\index{metric Levi-Civita connection} is obtained from the dual $\nabla^{(\alpha)}$ and $\nabla^{(-\alpha)}$ $\alpha$-connections:

\begin{equation}
\nabla^{(0)}=\frac{ \nabla^{(\alpha)} + \nabla^{(-\alpha)} }{2}.
\end{equation}

In particular, the $ \nabla^{(1)}$ connection is called the exponential connection\index{exponential connection}
and the $ \nabla^{(-1)}$ connection is called the mixture connection\index{mixture connection}.
The mixture/exponential geodesics for two distributions $p(x)$ and $q(x)$ induced by these dual connections are defined by

\begin{equation}
\gamma_\lambda(x)=(1-\lambda)p(x)+\lambda q(x)
\end{equation}

\begin{equation}
\log\gamma_\lambda(x)=(1-\lambda) \log p(x)+ \lambda\log q(x) -\log Z_\lambda,
\end{equation}
with $Z_\lambda$ the normalization coefficient $Z_\lambda=\int f^{(1-\lambda)}(x)g^\lambda(x)\dx$.
The exponential connection can be equivalently rewritten as:

\begin{equation}
\gamma_\lambda(x)=\frac{1}{Z_\lambda} p(x)^{(1-\lambda)}+ aq(x)^{\lambda}
\end{equation}
The dual connections are also called conjugate connections.\index{conjugate connection}

The canonical divergence\index{canonical divergence} on exponential family distributions  of these dually flat statistical manifolds  is shown to be:

\begin{equation}
D(p(x;\theta_1) || p(x;\theta_2)=F(\theta_1)+F^*(\theta_2)-<\theta_1,\eta_2>
\end{equation}

This canonical divergence\index{canonical divergence} can be rewritten as a Bregman divergence on the natural parameter space:

\begin{equation}
D(p(x;\theta_1) || p(x;\theta_2))=B_F(\theta_1||\theta_2)=F(\theta_1)-F(\theta_2)-<\theta_1-\theta_2,\grad F(\theta_2)>
\end{equation}

Furthermore, we have $B_F(\theta_1||\theta_2)=B_{F^*}(\eta_2||\eta_1)$, where $F^*$ is the Legendre conjugate of $F$, and $\eta_i=\grad F(\theta_i)$.

The Kullback-Leibler divergence\index{Kullback-Leibler} on two members of the same exponential family is equivalent to the Bregman divergence of the associated log-normalizer on swapped natural parameters:

\begin{equation}
\KL(p(x;\theta_1) || p(x;\theta_2))=\int p(x;\theta_1)\log \frac{p(x;\theta_1)}{p(x;\theta_2)} \dx=B_F(\theta_2||\theta_1)
\end{equation}

Note that the Bregman divergence may also be interpreted as a generalized relative entropy.\index{generalized relative entropy}
Indeed, 
\begin{equation}
KL(p||q)=H^\times(p||q)-H(p),
\end{equation}
 with 
\begin{equation}
H(p)=\int p(x)\log \frac{1}{p(x)} \dx
\end{equation}
 the Shannon entropy, and the cross-entropy $H^\times(p||q)\geq H(p)$:
 \begin{equation}
H^\times(p||q)=\int p(x)\log \frac{1}{q(x)} \dx
\end{equation}

Indeed, let $H_F(p)=-F(p)$ denote the generalized entropy, and 
\begin{equation}
H_F^\times (p||q)=-F(q)-<p-q,\grad F(q)>
\end{equation}
 the generalized cross-entropy.
The Bregman divergence can be considered as a generalized relative entropy:

\begin{equation}
B_F(p||q)=  H_F^\times (p||q) - H_F(p) 
\end{equation}

Bregman divergences are the canonical divergences of dually flat Riemannian manifolds~\cite{amari-2000}.
Bregman divergences extend naturally the quadratic distances (squared Euclidean and squared Mahalanobis distances), and generalize the notions of orthogonality,  projection, and Pythagoras' theorem.

The Bregman projection $q^\perp$ of a point $q$ onto a subspace $\W$ is the unique minimizer of $B_F(w||q)$ for $w\in\W$:

\begin{equation}
q^\perp = \arg\min B_F(w||q)
\end{equation}
This is a right-side projection. The left-side projection is simply a right-side projection for the Legendre conjugate generator $F^*$.

The following remarkable $3$-point property hold for any arbitrary Bregman divergence:

\begin{equation}
B_F(p||q)+B_F(q||r)=B_F(p||r)+ <p-q, \grad F(r)-\grad F(q)>.
\end{equation}
This formula may be interpreted as the law of generalized cosines.\index{law of generalized cosines}

The proof follows by mathematical rewriting:
{\small
\begin{eqnarray*}
B_F(p||q)+B_F(q||r) & = & F(p)-F(q)-<p-q,\nabla F(q)>+F(q)-F(r)-<q-r,\nabla F(r)>\\
&= & B_F(p||r)+<p-r,\nabla F(r)>+<r,\nabla F(r)>-<p,\nabla F(q)>+<q,\nabla
F(q)-\nabla F(r)>\\
&= & B_F(p||r)+<p-q, \nabla F(r)-\nabla F(q)>
\end{eqnarray*}
}

Geodesic $(pq)$ is orthogonal to dual geodesic $(rq)$.

Indeed, choosing $r=q^\perp$, we end-up with a generalized Pythagoras' theorem:

\begin{equation}
B_F(p||q)+B_F(q||r)=B_F(p||r)
\end{equation}
That is, triangle $p, q, r$ is a ``right-angle'' triangle.
The $\nabla$-geodesic $(pq)$ is orthogonal to the dual $\nabla^*$-geodesic $(qr)$.
The inner product $<p-q, \grad F(r)-\grad F(q)>$ vanishes. Note that the notion of orthogonality is not commutative.
Euclidean geometry is the special case of self-dual flat spaces with commutative orthogonality obtained for $F(x)=\frac{1}{2} x^T x$.

Banerjee et al.~\cite{banerjee-2005} formally proved the duality between exponential families and Bregman divergences for regular exponential families using Legendre transform:

\begin{equation}
\log p_F(x;\theta)=<t(x),\theta>-F(\theta)+k(x)=-B_{F*}(t(x)||\nabla F(\theta))+ \underbrace{F^*(x)+k(x)}_{=k_F(x)}
\end{equation}

This duality reveals key for designing an expectation-maximization algorithm using soft Bregman clustering.
The proof further reveals that

\begin{equation}
\X_F \subseteq \dom F^* .
\end{equation}
That is, the space of observations for the exponential families with log normalizer $F$ is included in the domain of the Legendre conjugate function.

For two very close points $p$ and $q\rightarrow p$, the Kullback-Leibler divergence is related to the Fisher metric by

\begin{equation}
\KL(p||q)\simeq \frac{1}{2} I^2(p,q)
\end{equation}

For exponential families, we can thus easily recover the Fisher information metric from the corresponding derivatives of the Bregman divergence.

However, it is difficult to compute Rao's distance\index{Rao's distance} $\int \sqrt{\dtheta^T\nabla^2 F(\theta) \dtheta} $ since it requires to compute the anti-derivative of $\sqrt{\nabla^2 F(\theta)}$. See for example, the work~\cite{raogamma-2003} that computes numerically an approximation of the Rao's distance for gamma distributions.

\subsection{Statistical inference}
\subsubsection{Maximum likelihood estimator}
Given $n$ i.i.d. observations $x_1, ..., x_n$ sampled from a given exponential family $p_F(x;\theta)$, the maximum likelihood estimator recover the parameter of the distribution by maximizing

\begin{equation}
\hat\theta=\argmax_{\theta} \prod_{i=1}^n p_F(x;\theta) .
\end{equation} 

It follows that the maximum likelihood estimator can be obtained from the center of mass of the sufficient statistics (the observed point\index{observed point}):\index{observed point}

\begin{equation}
\hat\theta=\grad F^*\left( \frac{1}{n} \sum_{i=1}^n t(x_i) \right)
\end{equation}

The variance of any unbiased estimator cannot beat the Cram\'er-Rao bound:\index{Cram\'er-Rao bound}

\begin{equation}
\var(\hat\theta)\geq I^{-1}(\theta)
\end{equation}

\subsubsection{Bayesian inference and conjugate priors}

The family of conjugate priors of an exponential family with likelihood function:

\begin{equation}
L(\theta; x_1, ..., x_n)= \exp \left( <\theta,\sum_{i=1}^n t(x_i)> -nF(\theta) + \sum_{i=1}^n k(x_i) \right)
\end{equation}
is

\begin{equation}
\exp\left( <\theta,g>-vF(\theta)   \right).
\end{equation}

\subsubsection{Statistical distances for exponential families}

Exponential families admit simple expressions for the entropies and related divergences~\cite{NN11}.
We summarize below the formula expressions, starting with R\'enyi $D_\alpha^R$ and Tsallis $D_\alpha^T(p:q)$ divergences (relative entropies):

\begin{eqnarray}
D_\alpha^R(p:q) &=& \frac{J_{F,\alpha}(\theta:\theta')}{1-\alpha}, \\
D_\alpha^T(p:q) & = &  \frac{e^{-J_{F,\alpha}}(\theta:\theta')-1}{\alpha-1}, \\
\KL(p:q) & = & \lim_{\alpha\rightarrow 1} D_\alpha^R(p:q) =  \lim_{\alpha\rightarrow 1} D_\alpha^T(p:q)= B_F(\theta' : \theta),
\end{eqnarray}
where 
\begin{equation}
J_{F,\alpha}(\theta:\theta')=\alpha F(\theta)+(1-\alpha) F(\theta')- F(\alpha\theta+(1-\alpha)\theta')=J_{F,1-\alpha}(\theta':\theta)
\end{equation}
is the skew Jensen divergence. 
Since the R\'enyi divergence for $\alpha=\frac{1}{2}$ is related to the Bhattacharrya coefficient $B(p,q)$ and Hellinger distance $H(p,q)$, this also yields closed-form expressions for members of the same exponential family:

\begin{eqnarray}
B(p,q) &=& e^{-J_{F,\frac{1}{2}}(\theta,\theta')}, \\
H(p,q) &=& \sqrt{1-e^{-J_{F,\frac{1}{2}}(\theta,\theta')}}.
\end{eqnarray}

R\'enyi $H^R_\alpha(p_F(x;\theta))$ and Tsallis entropies $H^T_\alpha(p_F(x;\theta))$ , including Shannon entropy $H(p_F(x;\theta))$ in the limit case, can be expressed respectively as 

\begin{eqnarray}
H^R_\alpha(p_F(x;\theta)) &=&  \frac{1}{1-\alpha} \left(F(\alpha\theta)-\alpha F(\theta)+\log E_p[e^{(\alpha-1)k(x)}]\right)   \\
H^T_\alpha(p_F(x;\theta)) &=&   \frac{1}{1-\alpha} \left((e^{F(\alpha\theta)-\alpha F(\theta)}) E_p[e^{(\alpha-1)k(x)}]-1  \right) \\
H(p_F(x;\theta))  &= &  F(\theta)-\inner{\theta}{\nabla F(\theta)}-E_p[k(x)]
\end{eqnarray}
The Shannon cross-entropy $H^\times(p_F(x;\theta) :p_F(x;\theta') )$ is given by
\begin{equation}
H^\times(p_F(x;\theta) :p_F(x;\theta') ) = F(\theta')-\inner{\theta'}{\nabla F(\theta)}  -E_p[k(x)]
\end{equation}

Thus these entropies admit closed-form formula  whenever the normalizing carrier measure is zero ($k(x)=0$):
\begin{eqnarray}
H^R_\alpha(p_F(x;\theta)) &=&  \frac{1}{1-\alpha} \left(F(\alpha\theta)-\alpha F(\theta) \right)   \\
H^T_\alpha(p_F(x;\theta)) &=&   \frac{1}{1-\alpha} \left(e^{F(\alpha\theta)-\alpha F(\theta)} -1  \right) \\
H(p_F(x;\theta))  &= &  F(\theta)-\inner{\theta}{\nabla F(\theta)}
\end{eqnarray}

Note that statistical distances are invariant by a monotonic function and sufficient statistics re-parameterization, and thus are expressed only using the log-normalizer $F$ and the natural parameter $\theta$.

\subsubsection{Mixture of exponential families}

Consider a mixture of exponential families of $k$-components

\begin{equation}
p_F(x;\theta_1, ..., \theta_k)=\sum_{i=1}^k w_i p_F(x;\theta_i),
\end{equation}
with $\sum_{i=1}^k w_i=1$ and all $w_i\geq 0$. 
Mixture of exponential families include the Gaussian mixture models (GMMs), mixtures of Gamma distributions, mixture of zero-mean Laplacians, etc.
To get a random sample from a mixture, we first draw randomly a number in $[0,1]$ to select the component, and then draw a random sample from the selected exponential family member.  

To fit a mixture model to a set of $n$ independently and identically distributed (i.i.d.) observations $x_1, ..., x_n$, we use the general expectation-maximization procedure~\cite{EM-1977}.

\begin{description}
\item[Initialization.]
We first compute a Bregman hard clustering on the $n$ observations $x_1, ..., x_n$ to get a collection of $k$ clusters.
Let $n_i$ denote the number of points in the $i$th cluster, and let $x_{i(j)}$ denote the points of the cluster for $j=1, ..., n_i$.
For each cluster, we initialize the $w_i=\frac{n_i}{n}$ to the proportion of points inside the cluster, and estimate the expectation/natural parameters $\eta_i/\theta_i$ using the observed point.
That is, $\eta_i=\frac{1}{n_i} \sum_{j=1}^{n_i} t(x_{i(j)})$ and in the dual coordinate system: $\theta_i=\grad F^{-1}(\frac{1}{n_i} \sum_{j=1}^{n_i} t(x_{i(j)}))$.

\item[Expectation step.]

\begin{equation}
p(i,j) = \frac{ w_j \exp ( -\mathrm{D_G}(t(x_i)| | \eta_j)) ) \exp ( k(x_i) ) }{ \sum_{l=1}^n w_l \exp{-\mathrm{D_G}(t(x_i) || \eta_l))}  \exp ( k(x_i) ) }
\end{equation}
with the Bregman divergence for the Legendre conjugate $G=F^*$ defined as:
\begin{equation}
D_G(p||q) = G(p) - G(q) - \langle p-q , \nabla G (q) \rangle
\end{equation}
We simplify\footnote{This step is crucial since $G(t(x))$ may not be defined. Consider for example, the univariate Gaussian distribution.
We have $G(\eta)=-\frac{1}{2}\log(\eta_1^2-\eta_2)$ with $t_1(x)=x$ and $t_2(x)=x^2$. Thus $G(t(x))=\log (x^2-x^2)$ is not defined. } the terms $G(t(x_i))$ in the numerator/denominator to get   
\begin{equation}
p(i,j) = \frac{ w_j \exp ( G(\eta_j ) + \langle t(x_i)-\eta_j , \nabla G (\eta_j) \rangle ) }
{ \sum_{l=1}^n w_l \exp ( G( \eta_l ) + \langle t(x_i)-\eta_l , \nabla G (\eta_l) \rangle ) } 
\end{equation}

\item[Maximization step.]

\begin{align}
w_j = & \frac{1}{N} \sum_{i=1}^N p(i,j) \\
\eta_j = & \frac{\sum_{i=1}^N p(i,j) t(x_i)}{\sum_{i=1}^N p(i,j)}
\end{align}

\end{description}

Given two mixtures of exponential families, we can bound the relative entropy of these distributions using Jensen's inequality on the convex Kullback-Leibler divergence as follows:

\begin{equation}
\KL(\sum_{i=1}^k w_i p_F(x;\theta_i) || \sum_{i=1}^{k'} w_i' p_F(x;\theta_i') ) 
\leq \sum_{i=1}^k\sum_{j=1}^{k'} w_iw_j' \KL(p_F(x;\theta_i)|| p_F(x;\theta_j') = 
\sum_{i=1}^k\sum_{j=1}^{k'} w_iw_j'B_F(\theta_j'||\theta_i)
\end{equation}

This bound is far too crude to be useful in practice.

We may consider approximating the relative entropy by matching components of the mixture, and get the following approximation: 

\begin{equation}
\KL(f || g)=\sum_{i=1}^k w_i \min_j B_F(\theta'_j||\theta_i)+\log \frac{w_i}{w_i'}
\end{equation}

In practice, the unscented transform yields a better approximation to the Kullback-Leibler divergence.
We consider $2dk$ sigma points as follows: For each component of the first mixture $f$, we decompose the Hessian $\Sigma_\theta=\grad ^2F(\theta)$ into $2d$ points
$\sqrt{(d\Sigma_\theta)_k}$, such that $\sqrt{(d\Sigma_\theta)_k}$ denote $k$-th column of the matrix square root of $\Sigma_\theta$.
Then the Kullback-Leibler divergence is approximated at the $2dk$ sigma points by

\begin{equation}
\frac{1}{2d} \sum_{i=1}^k w_i \sum_{j=1}^{2d} \log g(x_{i,j}),
\end{equation}
where $g$ is the probability density of the second mixture.

Even if it is widely known that GMMs can approximate arbitrarily finely any smooth probability density function, it is in fact possible to model any smooth density with a single member of an exponential family by defining the scalar product $<\cdot,\cdot>_{\mathcal{H}}$ in a
reproducing kernel Hilbert space~\cite{rkhs-expfam-2006} (RKHS) $\mathcal{H}$. Thus exponential families in RKHSs are universal density estimators~\cite{expfamuniversal-2004}.

\section*{Software library}
The jMEF is a Java library implementing the hard/soft/hierarchical techniques for exponential families with respect to sided and symmetrized Bregman divergences~\cite{2009-BregmanCentroids-TIT}:\\
{
\begin{center}
\url{http://www.lix.polytechnique.fr/~nielsen/MEF/}
\end{center}
}

\part{Exponential families: Flash cards}
{
\centering
\index{Exponential family catalog!Gaussian!Univariate}\section{Univariate Gaussian distribution}

\begin{tabular}{|c|c|}
	\hline
	PDF expression  & $f(x;\mu,\sigma^2) = \frac{1}{ \sqrt{2\pi \sigma^2} } \exp \left( - \frac{(x-\mu)^2}{ 2 \sigma^2} \right) \mbox{ for } x \in \mathds{R}$ \\
	\hline
	\hline
	Kullback-Leibler divergence &
	$D_{\mathrm{KL}}(f_P\|f_Q) = \frac{1}{2} \left(  2 \log \frac{\sigma_Q}{\sigma_P} + \frac{\sigma_P^2}{\sigma_Q^2} + \frac{(\mu_Q-\mu_P)^2}{\sigma_Q^2} -1\right)$ \\
	\hline
	MLE & $\hat \mu=\frac{1}{n}\sum_{i=1}^n x_i \qquad \hat\sigma=\sqrt{\frac{1}{n}\sum_{i=1}^n (x_i-\hat\mu)^2}$
	\\
	\hline
	\hline
	Source parameters  & $\mathbf{\Lambda} = ( \mu , \sigma^2 ) \ \in \mathds{R}\times\mathds{R}^+$ \\
	\hline
	Natural parameters & $\mathbf{\Theta} = ( \theta_1 , \theta_2 ) \ \in \mathds{R}\times\mathds{R}^-$ \\
	\hline
	Expectation parameters & $\mathbf{H} = ( \eta_1 , \eta_2 ) \ \in \mathds{R}\times\mathds{R}^+$ \\
	\hline
	\hline
	$\mathbf{\Lambda} \rightarrow \mathbf{\Theta}$ &
	$\mathbf{\Theta} =	\left( \frac{\mu}{\sigma^2} , -\frac{1}{2\sigma^2} \right)$ \\
	\hline
	$\mathbf{\Theta} \rightarrow \mathbf{\Lambda}$ &
	$\mathbf{\Lambda} = \left( -\frac{\theta_1}{2 \theta_2} , -\frac{1}{2 \theta_2} \right) $  \\
	\hline
	$\mathbf{\Lambda} \rightarrow \mathbf{H}$ &
	$\mathbf{H} = \left( \mu , \sigma^2 + \mu^2 \right)$ \\
	\hline
	$\mathbf{H} \rightarrow \mathbf{\Lambda}$ &
	$\mathbf{\Lambda} = \left( \eta_1 , \eta_2 - \eta_1^2 \right) $  \\
	\hline
	$\mathbf{\Theta} \rightarrow \mathbf{H}$ &
	$\mathbf{H} = \nabla F( \mathbf{\Theta} ) $ \\
	\hline
	$\mathbf{H} \rightarrow \mathbf{\Theta}$ &
	$\mathbf{\Theta} = \nabla G( \mathbf{H} ) $  \\
	\hline
	\hline
	Log normalizer &
	$ F(\mathbf{\Theta}) = -\frac{\theta_1^2}{4\theta_2} + \frac{1}{2} \log \left( -\frac{\pi}{\theta_2} \right) $  \\
	\hline
	Gradient log normalizer & $
	\nabla F(\mathbf{\Theta}) = \left( -\frac{\theta_1}{2 \theta_2}  , -\frac{1}{2 \theta_2} + \frac{\theta_1^2}{4 \theta_2^2} \right) $  \\
	\hline
	G &
	$G(\mathbf{H}) = - \frac{1}{2} \log \left( \eta_1^2 - \eta_2 \right) + C $  \\
	\hline
	Gradient G &
	$\nabla G(\mathbf{H}) = \left( -\frac{\eta_1}{\eta_1^2-\eta_2} , \frac{1}{2 (\eta_1^2-\eta_2)} \right) $  \\
	\hline
	\hline
	Sufficient statistics & $ t(x) = (x , x^2) $  \\
	\hline
	Carrier measure & $ k(x) = 0 $  \\
	\hline
\end{tabular}
\index{Exponential family catalog!Gaussian!Univariate with fixed variance}\section{Univariate Gaussian distribution, $\sigma^2$ fixed}

\begin{tabular}{|c|c|}
	\hline
	PDF expression  &
	$f(x; \mu, \sigma^2) = \frac{1}{ \sqrt{2\pi \sigma^2} } \exp \left( - \frac{(x-\mu)^2}{ 2 \sigma^2} \right) \mbox{ for } x \in \mathds{R} $\\
	\hline
	\hline
	Kullback-Leibler divergence &
	$D_{\mathrm{KL}}(f_P\|f_Q) = \frac{(\mu_Q-\mu_P)^2}{2\sigma^2}$ \\
	\hline
	MLE & $\hat \mu=\frac{1}{n}\sum_{i=1}^n x_i $
	\\
	\hline
	\hline
	Source parameters  & $\mathbf{\Lambda} = \mu \ \in \mathds{R}$ \\
	\hline
	Natural parameters & $\mathbf{\Theta} = \theta  \ \in \mathds{R}$ \\
	\hline
	Expectation parameters & $\mathbf{H} = \eta \ \in \mathds{R}$ \\
	\hline
	\hline
	$\mathbf{\Lambda} \rightarrow \mathbf{\Theta}$ &
	$\mathbf{\Theta} =	\frac{\mu}{\sigma^2} $ \\
	\hline
	$\mathbf{\Theta} \rightarrow \mathbf{\Lambda}$ &
	$\mathbf{\Lambda} = \theta \sigma^2 $  \\
	\hline
	$\mathbf{\Lambda} \rightarrow \mathbf{H}$ &
	$\mathbf{H} = \mu $ \\
	\hline
	$\mathbf{H} \rightarrow \mathbf{\Lambda}$ &
	$\mathbf{\Lambda} = \eta $  \\
	\hline
	$\mathbf{\Theta} \rightarrow \mathbf{H}$ &
	$\mathbf{H} = \nabla F( \mathbf{\Theta} ) $ \\
	\hline
	$\mathbf{H} \rightarrow \mathbf{\Theta}$ &
	$\mathbf{\Theta} = \nabla G( \mathbf{H} ) $  \\
	\hline
	\hline
	Log normalizer & $
	F(\mathbf{\Theta})	= \frac{\sigma^2 \theta^2 +  \log(2 \pi \sigma^2)}{2} $ \\
	\hline
	Gradient log normalizer & $
	\nabla F( \mathbf{\Theta} ) = \sigma^2 \theta $ \\
	\hline
	G &
	$G(\mathbf{H}) = \frac{\eta^2}{2 \sigma^2} + C $  \\
	\hline
	Gradient G &
	$\nabla G( \mathbf{H} ) = \frac{\eta}{\sigma^2} $  \\
	\hline
	\hline
	Sufficient statistics & $ t(x) = x $  \\
	\hline
	Carrier measure & $ k(x) = -\frac{x^2}{2 \sigma^2} $  \\
	\hline
\end{tabular}

\index{Exponential family catalog!Gaussian!Multivariate}\section{Multivariate Gaussian distribution}

\begin{tabular}{|c|c|}
	\hline
	PDF expression  & $f(x;\mu,\Sigma) = \frac{1}{ (2\pi)^{d/2} |\Sigma|^{1/2} } \exp \left( - \frac{(x-\mu)^T \Sigma^{-1}(x-\mu)}{2} \right) \mbox{ for } x \in \mathds{R}^d $ \\
	\hline
	\hline
	Kullback-Leibler divergence &
	\begin{tabular}{rl}
	$D_{\mathrm{KL}}(f_P \| f_Q)$ & = $\frac{1}{2} \left( \log \left( \frac{\det \Sigma_Q}{\det \Sigma_P} \right)
	+ \mathrm{tr} \left( \Sigma_Q^{-1} \Sigma_P \right) \right) $ \\
	& $+ \frac{1}{2} \left( ( \mu_Q - \mu_P )^\top \Sigma_Q^{-1} ( \mu_Q - \mu_P ) - d \right)$
	\end{tabular}
	\\
	\hline
	MLE &  $\hat\mu=\frac{1}{n}\sum_{i=1}^n x_i \qquad \hat\Sigma= \frac{1}{n} \sum_{i=1}^n (x_i-\hat \mu) (x_i-\hat \mu)^T$
	\\
	\hline
	\hline
	Source parameters  & $\mathbf{\Lambda} = ( \mu , \Sigma ) \mbox{ with } \mu \in \mathds{R}^d \mbox{ and } \Sigma \succ 0  $ \\
	\hline
	Natural parameters & $\mathbf{\Theta} = ( \theta , \Theta )$ \\
	\hline
	Expectation parameters & $\mathbf{H} = ( \eta , H )$ \\
	\hline
	\hline
	$\mathbf{\Lambda} \rightarrow \mathbf{\Theta}$ & $\mathbf{\Theta} = \left( \Sigma^{-1} \mu , \frac{1}{2} \Sigma^{-1} \right)$ \\
	\hline
	$\mathbf{\Theta} \rightarrow \mathbf{\Lambda}$ & $ \mathbf{\Lambda} = \left( \frac{1}{2} \Theta^{-1} \theta ,  \frac{1}{2} \Theta^{-1} \right) $  \\
	\hline
	$\mathbf{\Lambda} \rightarrow \mathbf{H}$ & $\mathbf{H} = \left( \mu , - (\Sigma + \mu \mu^T) \right)$ \\
	\hline
	$\mathbf{H} \rightarrow \mathbf{\Lambda}$ & $ \mathbf{\Lambda} = \left( \eta , - (H + \eta \eta^T) \right) $  \\
	\hline
	$\mathbf{\Theta} \rightarrow \mathbf{H}$ & $ \mathbf{H} = \nabla F( \mathbf{\Theta} ) $  \\
	\hline
	$\mathbf{H} \rightarrow \mathbf{\Theta}$ & $ \mathbf{\Theta} = \nabla G( \mathbf{H} ) $  \\
	\hline
	\hline
	Log normalizer & $ F(\mathbf{\Theta})=\frac{1}{4} \mathrm{tr}(\Theta^{-1}\theta\theta^T) - \frac{1}{2} \log \det\Theta + \frac{d}{2} log \pi$  \\
	\hline
	Gradient log normalizer & $ \nabla F( \mathbf{\Theta} ) = \left( \frac{1}{2} \Theta^{-1} \theta , -\frac{1}{2} \Theta^{-1} -\frac{1}{4} (\Theta^{-1} \theta)(\Theta^{-1} \theta)^T \right) $  \\
	\hline
	G & $ G(\mathbf{H}) = - \frac{1}{2} \log \left( 1 + \eta^T H^{-1} \eta \right) - \frac{1}{2} \log \det (-H) - \frac{d}{2} \log (2 \pi e) $  \\
	\hline
	Gradient G & $ \nabla G(\mathbf{H}) = \left( -( H + \eta \eta^T )^{-1} \eta , -\frac{1}{2} ( H + \eta \eta^T )^{-1}  \right) $  \\
	\hline
	\hline
	Sufficient statistics & $ t(x) = (x , -x x^T) $  \\
	\hline
	Carrier measure & $ k(x) = 0 $  \\
	\hline
\end{tabular}

\index{Exponential family catalog!Gaussian!Multivariate isotropic}\section{Multivariate isotropic Gaussian distribution}

We assume identity variance-covariance matrix $\Sigma=\mathrm{Id}$.

\begin{tabular}{|c|c|}
	\hline
	PDF expression  & $f(x;\mu) = \frac{1}{ (2\pi)^{d/2} } \exp \left( - \frac{(x-\mu)^T (x-\mu)}{2} \right) \mbox{ for } x \in \mathds{R}^d $ \\
	\hline
	\hline
	Kullback-Leibler divergence &
	$D_{\mathrm{KL}}(f_P \| f_Q) = \frac{1}{2} ( \mu_Q - \mu_P )^\top( \mu_Q - \mu_P )$
	\\
	\hline
	MLE &  $\hat \mu=\frac{1}{n}\sum_{i=1}^n x_i$ 
	\\
	\hline
	\hline
	Source parameters  & $\mathbf{\Lambda} = \mu  \mbox{ with } \mu \in \mathds{R}^d $\\
	\hline
	Natural parameters & $\mathbf{\Theta} = \theta$ \\
	\hline
	Expectation parameters & $\mathbf{H} = \eta$ \\
	\hline
	\hline
	$\mathbf{\Lambda} \rightarrow \mathbf{\Theta}$ & $\mathbf{\theta} = \mu $ \\
	\hline
	$\mathbf{\Theta} \rightarrow \mathbf{\Lambda}$ & $ \mathbf{\Lambda} = \theta $  \\
	\hline
	$\mathbf{\Lambda} \rightarrow \mathbf{H}$ & $\mathbf{H} = \mu $ \\
	\hline
	$\mathbf{H} \rightarrow \mathbf{\Lambda}$ & $ \mathbf{\Lambda} = \eta $  \\
	\hline
	$\mathbf{\Theta} \rightarrow \mathbf{H}$ & $ \mathbf{H} = \nabla F( \mathbf{\Theta} ) $  \\
	\hline
	$\mathbf{H} \rightarrow \mathbf{\Theta}$ & $ \mathbf{\Theta} = \nabla G( \mathbf{H} ) $  \\
	\hline
	\hline
	Log normalizer & $ F(\mathbf{\theta}) = \frac{1}{2} \theta^\top\theta + \frac{d}{2}\log 2\pi$  \\
	\hline
	Gradient log normalizer & $\nabla F(\mathbf{\theta}) = \theta$  \\
	\hline
	G & $ G(\mathbf{\eta}) =  F(\mathbf{\theta})= \frac{1}{2} \eta^\top\eta + \frac{d}{2}\log 2\pi$ \\
	\hline
	Gradient G & $ \nabla G(\mathbf{\eta}) = \eta$  \\
	\hline
	\hline
	Sufficient statistics & $ t(x) = x $  \\
	\hline
	Carrier measure & $ k(x) = -\frac{1}{2}x^\top x $  \\
	\hline
\end{tabular}

\index{Exponential family catalog!Poisson}\section{Poisson distribution}

\begin{tabular}{|c|c|}
	\hline
	PDF expression  &
	$f( x ; \lambda ) = \frac{\lambda^x \exp(-\lambda)}{x!} \mbox{ for }	 x \in \mathds{N}^+$ \\
	\hline
	\hline
	Kullback-Leibler divergence &
	$ D_{\mathrm{KL}}(f_P\|f_Q) = \lambda_Q - \lambda_P \left( 1 + \log \left( \frac{\lambda_Q}{\lambda_P} \right) \right) $\\
	\hline
	MLE &  $\hat\lambda=\frac{1}{n} \sum_{i=1}^n x_i$
	\\
	\hline
	\hline
	Source parameters  & $\mathbf{\Lambda} = \lambda \ \in \mathds{R}^+ $\\
	\hline
	Natural parameters & $\mathbf{\Theta} = \theta \ \in \mathds{R} $ \\
	\hline
	Expectation parameters & $\mathbf{H} = \eta \ \in \mathds{R}^+ $ \\
	\hline
	\hline
	$\mathbf{\Lambda} \rightarrow \mathbf{\Theta}$ &
	$\mathbf{\Theta} =	\log \lambda $ \\
	\hline
	$\mathbf{\Theta} \rightarrow \mathbf{\Lambda}$ &
	$\mathbf{\Lambda} = \exp \theta $ \\
	\hline
	$\mathbf{\Lambda} \rightarrow \mathbf{H}$ &
	$\mathbf{H} = \lambda $ \\
	\hline
	$\mathbf{H} \rightarrow \mathbf{\Lambda}$ &
	$\mathbf{\Lambda} = \eta$  \\
	\hline
	$\mathbf{\Theta} \rightarrow \mathbf{H}$ &
	$\mathbf{H} = \nabla F( \mathbf{\Theta} ) $ \\
	\hline
	$\mathbf{H} \rightarrow \mathbf{\Theta}$ &
	$\mathbf{\Theta} = \nabla G( \mathbf{H} ) $ \\
	\hline
	\hline
	Log normalizer & $
	F(\mathbf{\Theta})	= \exp \theta $  \\
	\hline
	Gradient log normalizer & $
	\nabla F( \mathbf{\Theta} ) = \exp \theta $  \\
	\hline
	G &
	$G(\mathbf{H}) = \eta \log \eta - \eta + C$  \\
	\hline
	Gradient G &
	$\nabla G( \mathbf{H} ) = \log \eta $  \\
	\hline
	\hline
	Sufficient statistics & $ t(x) = x $  \\
	\hline
	Carrier measure & $ k(x) = - \log (x!) $  \\
	\hline
\end{tabular}
\index{Exponential family catalog!Laplacian}\section{Centered Laplacian distribution, $\mu=0$}

\begin{tabular}{|c|c|}
	\hline
	PDF expression  & $f(x;\sigma) = \frac{1}{ 2 \sigma } \exp \left( - \frac{|x|}{\sigma} \right) \mbox{ for } x \in \mathds{R} $ \\
	\hline
	\hline
	Kullback-Leibler divergence &
	$ D_{\mathrm{KL}}(f_P\|f_Q) = \log \left( \frac{\sigma_Q}{\sigma_P} \right) + \frac{\sigma_P - \sigma_Q}{\sigma_Q}$ \\
	\hline
	MLE & $\hat\sigma=\frac{1}{n} \sum_{i=1}^n |x_i|$
	\\
	\hline
	\hline
	Source parameters  & $\mathbf{\Lambda} = \sigma \ \in \mathds{R}^+$ \\
	\hline
	Natural parameters & $\mathbf{\Theta} = \theta \ \in \mathds{R}^-$ \\
	\hline
	Expectation parameters & $\mathbf{H} = \eta \ \in \mathds{R}^+$ \\
	\hline
	\hline
	$\mathbf{\Lambda} \rightarrow \mathbf{\Theta}$ & $ \mathbf{\Theta} = -\frac{1}{\sigma} $ \\
	\hline
	$\mathbf{\Theta} \rightarrow \mathbf{\Lambda}$ & $ \mathbf{\Lambda} = -\frac{1}{\theta} $  \\
	\hline
	$\mathbf{\Lambda} \rightarrow \mathbf{H}$ &	$\mathbf{H} = \sigma $ \\
	\hline
	$\mathbf{H} \rightarrow \mathbf{\Lambda}$ &	$\mathbf{\Lambda} = \eta$  \\
	\hline
	$\mathbf{\Theta} \rightarrow \mathbf{H}$ & $ \mathbf{H} = \nabla F( \mathbf{\Theta} ) $  \\
	\hline
	$\mathbf{H} \rightarrow \mathbf{\Theta}$ & $ \mathbf{\Theta} = \nabla G( \mathbf{H} ) $  \\
	\hline
	\hline
	Log normalizer & $ F(\mathbf{\Theta}) = \log \left( -\frac{2}{\theta} \right)$  \\
	\hline
	Gradient log normalizer & $ \nabla F( \mathbf{\Theta} ) = -\frac{1}{\theta} $  \\
	\hline
	G & $ G(\mathbf{H}) = - \log \eta + C $  \\
	\hline
	Gradient G & $ \nabla G(\mathbf{H}) = -\frac{1}{\eta} $ \\
	\hline
	\hline
	Sufficient statistics & $ t(x) = |x| $  \\
	\hline
	Carrier measure & $ k(x) = 0 $  \\
	\hline
\end{tabular}
\index{Exponential family catalog!Bernoulli}\section{Bernoulli distribution}

\begin{tabular}{|c|c|}
	\hline
	PDF expression  &
	$f( x ; p ) = p^x (1-p)^{1-x} \mbox{ for } x \in \{0,1\}

	$ \\
	\hline
	\hline
	Kullback-Leibler divergence &
	$ D_{\mathrm{KL}}(f_1\|f_2) = \log \left( \frac{1-p_1}{1-p_2} \right) - p_1 \log \left( \frac{p_2(1-p_1)}{p_1(1-p_2)} \right)$ \\
	\hline
	MLE & $\hat p=\frac{1}{n}\sum_{i=1}^n x_i$
	\\
	\hline
	\hline
	Source parameters  & $\mathbf{\Lambda} = p \in [0,1]$ \\
	\hline
	Natural parameters & $\mathbf{\Theta} = \theta \in \mathds{R}^+$ \\
	\hline
	Expectation parameters & $\mathbf{H} = \eta \in [0,1]$ \\
	\hline
	\hline
	$\mathbf{\Lambda} \rightarrow \mathbf{\Theta}$ &
	$\mathbf{\Theta} =	\log \left( \frac{p}{1-p} \right) $ \\
	\hline
	$\mathbf{\Theta} \rightarrow \mathbf{\Lambda}$ &
	$\mathbf{\Lambda} = \frac{\exp\theta}{1+\exp\theta} $ \\
	\hline
	$\mathbf{\Lambda} \rightarrow \mathbf{H}$ &
	$\mathbf{H} = p$ \\
	\hline
	$\mathbf{H} \rightarrow \mathbf{\Lambda}$ &
	$\mathbf{\Lambda} = \eta$  \\
	\hline
	$\mathbf{\Theta} \rightarrow \mathbf{H}$ &
	$\mathbf{H} = \nabla F( \mathbf{\Theta} ) $ \\
	\hline
	$\mathbf{H} \rightarrow \mathbf{\Theta}$ &
	$\mathbf{\Theta} = \nabla G( \mathbf{H} ) $ \\
	\hline
	\hline
	Log normalizer & $
	F(\mathbf{\Theta})	= \log \left( 1 + \exp \theta \right) $  \\
	\hline
	Gradient log normalizer & $
	\nabla F( \mathbf{\Theta} ) = \frac{\exp \theta}{1 + \exp \theta} $  \\
	\hline
	G &
	$G(\mathbf{H}) = \log \left( \frac{\eta}{1-\eta} \right) \eta - \log \left( \frac{1}{1-\eta} \right) + C$  \\
	\hline
	Gradient G &
	$\nabla G( \mathbf{H} ) = \log \left( \frac{\eta}{1-\eta} \right)$  \\
	\hline
	\hline
	Sufficient statistics & $ t(x) = x$  \\
	\hline
	Carrier measure & $ k(x) = 0$  \\
	\hline
\end{tabular}
\index{Exponential family catalog!Binomial}\section{Binomial distribution, $n$ fixed $\in \mathds{N}^+$}

\begin{tabular}{|c|c|}
	\hline
	PDF expression  &
	$f( x ; n, p ) = \frac{n!}{x!(n-x)!} p^x (1-p)^{n-x} $ where $x \in \mathds{N}^+$\\
	\hline
	\hline
	Kullback-Leibler divergence &
	$ D_{\mathrm{KL}}(f_1\|f_2) = n (1-p_1) \log \left( \frac{1-p_1}{1-p_2} \right) + n p_1 \log \left( \frac{p_1}{p_2} \right) $ \\
	\hline
	MLE & $\hat p =\frac{1}{n}\sum_{i=1}^n x_i$
	\\
	\hline
	\hline
	Source parameters  & $\mathbf{\Lambda} = p \ \in [0,1]$ \\
	\hline
	Natural parameters & $\mathbf{\Theta} = \theta \ \in \mathds{R}$ \\
	\hline
	Expectation parameters & $\mathbf{H} = \eta \ \in \mathds{R}^+ $ \\
	\hline
	\hline
	$\mathbf{\Lambda} \rightarrow \mathbf{\Theta}$ &
	$\mathbf{\Theta} =	\log \left( \frac{p}{1-p} \right) $ \\
	\hline
	$\mathbf{\Theta} \rightarrow \mathbf{\Lambda}$ &
	$\mathbf{\Lambda} = \frac{\exp \theta}{1 + \exp \theta} $ \\
	\hline
	$\mathbf{\Lambda} \rightarrow \mathbf{H}$ &
	$\mathbf{H} = np $ \\
	\hline
	$\mathbf{H} \rightarrow \mathbf{\Lambda}$ &
	$\mathbf{\Lambda} = \frac{\eta}{n} $  \\
	\hline
	$\mathbf{\Theta} \rightarrow \mathbf{H}$ &
	$\mathbf{H} = \nabla F( \mathbf{\Theta} ) $ \\
	\hline
	$\mathbf{H} \rightarrow \mathbf{\Theta}$ &
	$\mathbf{\Theta} = \nabla G( \mathbf{H} ) $ \\
	\hline
	\hline
	Log normalizer & $
	F(\mathbf{\Theta})	= n \log (1 + \exp \theta) - \log (n!) $  \\
	\hline
	Gradient log normalizer & $
	\nabla F( \mathbf{\Theta} ) = \frac{n \exp \theta}{1 + \exp \theta} $  \\
	\hline
	G &
	$G(\mathbf{H}) = \eta \log \left( \frac{\eta}{n-\eta} \right)  - n \log\left( \frac{n}{n-\eta} \right) + C$  \\
	\hline
	Gradient G &
	$\nabla G( \mathbf{H} ) = \log \left( \frac{\eta}{n-\eta} \right) $  \\
	\hline
	\hline
	Sufficient statistics & $ t(x) = x $  \\
	\hline
	Carrier measure & $ k(x) = - \log (x! (n-x)!) $  \\
	\hline
\end{tabular}
\index{Exponential family catalog!Multinomial}\section{Multinomial distribution, $n$ fixed}

\begin{tabular}{|c|c|}
	\hline
	PDF expression  & $f(x_1,\cdots,x_k;p_1,\cdots,p_k,n) = \frac{n!}{x_1! \cdots x_k!} p_1^{x_1} \cdots p_k^{x_k}  \mbox{ for } x_i \in \mathds{N}^+$ \\
	\hline
	\hline
	Kullback-Leibler divergence &
	$D_{\mathrm{KL}}(f_\alpha\|f_\beta) = n p_{\alpha,k} \log \frac{p_{\alpha,k}}{p_{\beta,k}} - n \sum_{i=1}^{k-1} p_{\alpha,i} \log \frac{p_{\beta,i}}{p_{\alpha,i}} $ \\
	\hline
	MLE &   $\hat p_i=\frac{n_i}{n}$
	\\
	\hline
	\hline
	Source parameters  & $\mathbf{\Lambda} = (p_1, \cdots, p_k) \ \in [0,1]^k \mbox{ with } \sum_i p_i = 1 $ \\
	\hline
	Natural parameters & $\mathbf{\Theta} = (\theta_1, \cdots, \theta_{k-1}) \ \in \mathds{R}^{k-1} $ \\
	\hline
	Expectation parameters & $\mathbf{H} = (\eta_1, \cdots, \eta_{k-1}) \ \in [0,n]^{k-1} $ \\
	\hline
	\hline
	$\mathbf{\Lambda} \rightarrow \mathbf{\Theta}$ &
	$\mathbf{\Theta} = \left( \log \left( \frac{p_i}{p_k} \right) \right)_i$ \\
	\hline
	$\mathbf{\Theta} \rightarrow \mathbf{\Lambda}$ &
	$\mathbf{\Lambda} =  
	\begin{cases}
	p_i = \frac{\exp \theta_i}{1 + \sum_{j=1}^{k-1} \exp \theta_j} & \mbox{if $i<k$}\\
	p_k = \frac{1}{1 + \sum_{j=1}^{k-1} \exp \theta_j }
	\end{cases}$\\
	\hline
	$\mathbf{\Lambda} \rightarrow \mathbf{H}$ &
	$\mathbf{H} = \left( n p_i \right)_i$ \\
	\hline
	$\mathbf{H} \rightarrow \mathbf{\Lambda}$ &
	$\mathbf{\Lambda} =   
	\begin{cases}
	p_i = \frac{\eta_i}{n} & \mbox{if $i<k$}\\
	p_k = \frac{n - \sum_{j=1}^{k-1} \eta_j}{n}
	\end{cases}$\\
	\hline
	$\mathbf{\Theta} \rightarrow \mathbf{H}$ &
	$\mathbf{H} = \nabla F( \mathbf{\Theta} ) $ \\
	\hline
	$\mathbf{H} \rightarrow \mathbf{\Theta}$ &
	$\mathbf{\Theta} = \nabla G( \mathbf{H} ) $ \\
	\hline
	\hline
	Log normalizer &
	$ F(\mathbf{\Theta}) = n \log \left( 1 + \sum_{i=1}^{k-1} \exp \theta_i \right) - \log n!$  \\
	\hline
	Gradient log normalizer & $
	\nabla F(\mathbf{\Theta}) = \left( \frac{n \exp \theta_i}{1 + \sum_{j=1}^{k-1} \exp \theta_j} \right)_i $  \\
	\hline
	G &
	$G(\mathbf{H}) = \left( \sum_{i=1}^{k-1} \eta_i \log \eta_i \right) + \left( n - \sum_{i=1}^{k-1} \eta_i \right) \log \left( n - \sum_{i=1}^{k-1} \eta_i \right) + C$  \\
	\hline
	Gradient G &
	$\nabla G(\mathbf{H}) = \left( \log \left( \frac{\eta_i}{n - \sum_{j=1}^{k-1} \eta_j} \right) \right)_i$  \\
	\hline
	\hline
	Sufficient statistics & $ t(x) = (x_1, \cdots, x_{k-1})$  \\
	\hline
	Carrier measure & $ k(x) = - \sum_{i=1}^{k} \log x_i !$  \\
	\hline
\end{tabular}
\index{Exponential family catalog!Rayleigh}

\section{Rayleigh distribution}

\begin{tabular}{|c|c|}
	\hline
	PDF expression  & $f(x;\sigma^2) = \frac{x}{\sigma^2} \exp \left( -\frac{x^2}{2\sigma^2} \right) $ \\
	\hline
	\hline
	Kullback-Leibler divergence &

	$D_{\mathrm{KL}}(f_P \| f_Q) = \log \left( \frac{\sigma_Q^2}{\sigma_P^2} \right) + \frac{ \sigma_P^2 - \sigma_Q^2 }{\sigma_Q^2}$

	\\
	\hline
	MLE &  $\hat\sigma=\sqrt{\frac{1}{2n}\sum_{i=1}^n x_i^2}$
	\\
	\hline
	\hline
	Source parameters  & $\mathbf{\Lambda} = \sigma^2  \in \mathds{R}^+ $\\
	\hline
	Natural parameters & $\mathbf{\Theta} = \theta \in \mathds{R}^- $ \\
	\hline
	Expectation parameters & $\mathbf{H} = \eta  \in \mathds{R}^+ $ \\
	\hline
	\hline
	$\mathbf{\Lambda} \rightarrow \mathbf{\Theta}$ & $\mathbf{\Theta} = -\frac{1}{2 \sigma^2} $ \\
	\hline
	$\mathbf{\Theta} \rightarrow \mathbf{\Lambda}$ & $ \mathbf{\Lambda} = -\frac{1}{2\theta}$  \\
	\hline
	$\mathbf{\Lambda} \rightarrow \mathbf{H}$ & $\mathbf{H} = 2 \sigma^2$ \\
	\hline
	$\mathbf{H} \rightarrow \mathbf{\Lambda}$ & $ \mathbf{\Lambda} = \frac{\eta}{2} $  \\
	\hline
	$\mathbf{\Theta} \rightarrow \mathbf{H}$ & $ \mathbf{H} = \nabla F( \mathbf{\Theta} ) $  \\
	\hline
	$\mathbf{H} \rightarrow \mathbf{\Theta}$ & $ \mathbf{\Theta} = \nabla G( \mathbf{H} ) $  \\
	\hline
	\hline
	Log normalizer & $ F(\mathbf{\theta}) = - \log (-2 \theta)$  \\
	\hline
	Gradient log normalizer & $ \nabla F(\mathbf{\theta}) = -\frac{1}{\theta}$  \\
	\hline
	G & $ G(\mathbf{\eta}) = - \log \eta $   \\
	\hline
	Gradient G & $ \nabla G(\mathbf{\eta}) = -\frac{1}{\eta}$  \\
	\hline
	\hline
	Sufficient statistics & $ t(x) = x^2 $  \\
	\hline
	Carrier measure & $ k(x) = \log x $  \\
	\hline
\end{tabular}

\index{Exponential family catalog!Gamma}

\section{Gamma distribution}

\def\digamma{\Psi}

$\Gamma(k)=(k-1)!$ and $\digamma(x)=\frac{\Gamma'(x)}{\Gamma(x)}$

\begin{tabular}{|c|c|}
	\hline
	PDF expression  & $f(x;\lambda,k) = x^{k-1} \frac{e^{-\frac{x}{\lambda}}}{\lambda^k \Gamma(k)}$ \\
	\hline
	\hline
	Kullback-Leibler divergence &
	$D_{\mathrm{KL}}(f_P\|f_Q) = \log\frac{\Gamma(\lambda_Q)}{\lambda_P} + (k_P-k_Q)(\digamma(k_P)-\log\lambda_P)+k_P\frac{\lambda_Q-\lambda_P}{\lambda_P} $ \\
	\hline
	MLE & $\hat k\hat\lambda=\sum_{i=1}^n x_i$ \\
	    & $\log\hat k-\digamma(k)=\log (\frac{1}{n}\sum_{i=1}^n x_i) - \frac{1}{n}\sum_{i=1}^n\log x_i$
	\\
	\hline
	\hline
	Source parameters  & $\mathbf{\Lambda} = (\lambda,k)$ \\
	\hline
	Natural parameters & $\mathbf{\Theta} = (k-1,-\frac{1}{\lambda})$ \\
	\hline
	Expectation parameters & $\mathbf{H} = $ \\
	\hline
	\hline
	$\mathbf{\Lambda} \rightarrow \mathbf{\Theta}$ &
	$\mathbf{\Theta} = $ \\
	\hline
	$\mathbf{\Theta} \rightarrow \mathbf{\Lambda}$ &
	$\mathbf{\Lambda} =  $  \\
	\hline
	$\mathbf{\Lambda} \rightarrow \mathbf{H}$ &
	$\mathbf{H} =  $ \\
	\hline
	$\mathbf{H} \rightarrow \mathbf{\Lambda}$ &
	$\mathbf{\Lambda} = $  \\
	\hline
	$\mathbf{\Theta} \rightarrow \mathbf{H}$ &
	$\mathbf{H} = $ \\
	\hline
	$\mathbf{H} \rightarrow \mathbf{\Theta}$ &
	$\mathbf{\Theta} = $  \\
	\hline
	\hline
	Log normalizer &
	$ F(\mathbf{\Theta}) = \log\Gamma(\theta_1+1)+(\theta_1+1)\log \frac{-1}{\theta_2}$  \\
	\hline
	Gradient log normalizer & $
	\nabla F(\mathbf{\Theta}) =  (\digamma(\theta_1+1)+\log\frac{-1}{\theta_2} , -\frac{\theta_1+1}{\theta_2}) $  \\
	\hline
	G &
	$G(\mathbf{H})  $ non-closed form  \\
	\hline
	Gradient G &
	$\nabla G(\mathbf{H}) $ non-closed form  \\
	\hline
	\hline
	Sufficient statistics & $ t(x) = (x,\log x)$  \\
	\hline
	Carrier measure & $ k(x) = $  \\
	\hline
\end{tabular}
\index{Exponential family catalog!Gamma}\section{Beta distributions}
\index{Exponential family catalog!Beta}

$\Gamma(k)=(k-1)!$, $B(\alpha,\beta)=\frac{\Gamma(\alpha)\Gamma(\beta)}{\Gamma(\alpha+\beta)}$ and  $\digamma(x)=\frac{\Gamma'(x)}{\Gamma(x)}$
\begin{tabular}{|c|c|}
	\hline
	PDF expression  & $f(x;\alpha,\beta) = \frac{1}{B(\alpha,\beta} x^{\alpha-1} (1-x)^{\beta-1} $ \\
	\hline
	\hline
	Kullback-Leibler divergence & \parbox{10cm}{
	$D_{\mathrm{KL}}(f_P\|f_Q) = \log\frac{B(\alpha_Q,\beta_Q)}{B(\alpha_P,\beta_P)}-(\beta_Q-\beta_P)\digamma(\alpha_P)
	-(\beta_Q-\beta_P)\digamma(\beta_P)+(\alpha_Q-\alpha_P+\beta_Q-\beta_P)\digamma(\alpha_P+\alpha_Q) $ }
	\\
	\hline
	MLE &  $\left\{  \begin{array}{l} \frac{\partial B(\alpha,\beta)}{\partial\alpha}=B(\alpha,\beta) \sum_{i=1}^n \log x_i \\
	\frac{\partial B(\alpha,\beta)}{\partial\beta}=B(\alpha,\beta)\sum_{i=1}^n \log (1-x_i) 
	  \end{array} \right. $
	\\
	\hline
	\hline
	Source parameters  & $\mathbf{\Lambda} = (\alpha,\beta) $ \\
	\hline
	Natural parameters & $\mathbf{\Theta} = $ \\
	\hline
	Expectation parameters & $\mathbf{H} = $ \\
	\hline
	\hline
	$\mathbf{\Lambda} \rightarrow \mathbf{\Theta}$ &
	$\mathbf{\Theta} = $ \\
	\hline
	$\mathbf{\Theta} \rightarrow \mathbf{\Lambda}$ &
	$\mathbf{\Lambda} =  $  \\
	\hline
	$\mathbf{\Lambda} \rightarrow \mathbf{H}$ &
	$\mathbf{H} =  $ \\
	\hline
	$\mathbf{H} \rightarrow \mathbf{\Lambda}$ &
	$\mathbf{\Lambda} = $  \\
	\hline
	$\mathbf{\Theta} \rightarrow \mathbf{H}$ &
	$\mathbf{H} = $ \\
	\hline
	$\mathbf{H} \rightarrow \mathbf{\Theta}$ &
	$\mathbf{\Theta} = $  \\
	\hline
	\hline
	Log normalizer &
	$ F(\mathbf{\Theta}) = \log B(\theta_1+1,\theta_2+1)$  \\
	\hline
	Gradient log normalizer & $
	\nabla F(\mathbf{\Theta}) =  (\digamma(\theta_1+1)-\digamma(\theta_1+\theta_2+2), \digamma(\theta_2+1)-\digamma(\theta_1+\theta_2+2)) $  \\
	\hline
	G &
	$G(\mathbf{H}) $  non-closed form \\
	\hline
	Gradient G &
	$\nabla G(\mathbf{H}) $ non-closed form  \\
	\hline
	\hline
	Sufficient statistics & $ t(x) = (\log x, \log (1-x))$  \\
	\hline
	Carrier measure & $ k(x) = $  \\
	\hline
\end{tabular}
\index{Exponential family catalog!Pareto}\include{Pareto}
}


\printindex

\end{document}